\documentclass[letterpaper, 10 pt, conference]{ieeeconf}  

\IEEEoverridecommandlockouts                              

\usepackage{graphics} 
\usepackage{graphicx}
\usepackage{amsmath} 
\usepackage{amssymb}  
\usepackage{mathtools}

\usepackage{times}

\usepackage{multicol}
\usepackage[bookmarks=true]{hyperref}
\usepackage[colorinlistoftodos]{todonotes}
\DeclareMathOperator*{\argmax}{arg\,max}

\newcommand{\triangulation}{T}
\newcommand{\pixel}{x}
\newcommand{\pixels}{\mathbf{\pixel}}
\newcommand{\referencePixel}{\pixel^{ref}}

\newcommand{\segmentation}{S}
\newcommand{\segmentationPrior}{S_P}
\newcommand{\disparity}{d}

\newcommand{\allPriors}{\triangulation, \segmentationPrior, \referencePixel}

\newcommand{\disparityDistribution}{p(\disparity | \triangulation, \referencePixel)}
\newcommand{\pixelLikelihood}{p(\pixels|d,\referencePixel,\segmentation)}
\newcommand{\segmentationProbability}{p(\segmentation|\segmentationPrior)}

\newcommand{\generalEDistr}{p(Z|X,\theta^{old})}
\newcommand{\generalMDistr}{\log p(X,Z | \theta)}
\newcommand{\generalPDistr}{\log p(\theta)}
\newcommand{\generalEM}{\generalEDistr \generalMDistr}

\newcommand{\EDistr}{p(\segmentation | \pixels, \disparity^{old}, \allPriors)}
\newcommand{\MDistr}{\log p(\segmentation, \pixels | \disparity, \allPriors)}

\usepackage{cleveref}
\usepackage{changes}
\usepackage{todonotes}

\title{\LARGE \bf
Removing Dynamic Objects for Static Scene Reconstruction using Light Fields
}

\author{Pushyami Kaveti$^{1}$, Sammie Katt$^{1}$ and Hanumant Singh$^{2}$
\thanks{*This work was partly supported by ONR Grant N00014-19-1-2131}
\thanks{$^{1}$Khoury College of Computer Science, Northeastern University, Boston, MA (email: {\tt\small kaveti.p@husky.neu.edu, katt.s@husky.neu.edu})}%
\thanks{$^{2}$Department of Electrical and Computer Engineering, Northeastern University,Boston, MA (email: {\tt\small ha.singh@northeastern.edu})}%
}

\begin{document}


\maketitle
\thispagestyle{empty}
\pagestyle{empty}

\begin{abstract}

There is a general expectation that robots should operate in environments that consist of static and dynamic entities including people, furniture and automobiles. These dynamic environments pose challenges to visual simultaneous localization and mapping (SLAM) algorithms by introducing errors into the front-end. Light fields provide one possible method for addressing such problems by capturing a more complete visual information of a scene. In contrast to a single ray from a perspective camera, Light Fields capture a bundle of light rays emerging from a single point in space, allowing us to see through dynamic objects by refocusing past them. 

In this paper we present a method to synthesize a refocused image of the static background in the presence of dynamic objects that uses a light-field acquired with a linear camera array. We simultaneously estimate both the depth and the refocused image of the static scene using semantic segmentation for detecting dynamic objects in a single time step. This eliminates the need for initializing a static map . The algorithm is parallelizable and is implemented on GPU allowing us execute it at close to real time speeds. We demonstrate the effectiveness of our method on real-world data acquired using a small robot with a five camera array. 

\end{abstract}

\section{INTRODUCTION}

\begin{figure*}[ht!]
\includegraphics[clip,trim={0cm 1cm 0cm 0cm}, width=0.95\linewidth]{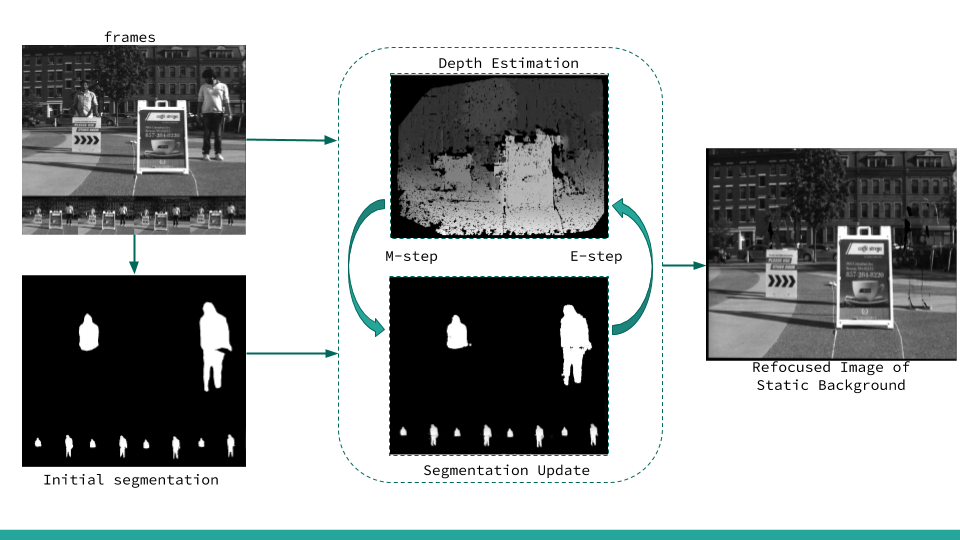}
\caption{Block diagram of the static background reconstruction pipeline. Frames from the linear camera array and the segmentation probabilities are given as input into the EM framework where we iteratively estimate the depth and segmentation maps. The optimized depth and segmentation is used to compute the refocused image.}
\label{fig:blk}
\end{figure*}
SLAM facilitates robot navigation and mapping and is one of the primary tasks in mobile robotics.
Most of the research in SLAM assumes static environments, but the real world is complex and dynamic, making practical applications (autonomous motion on a crowded road or in a corridor) difficult. Algorithms tend to fail in the presence of dynamic objects due to errors in feature matching, loop closure and pose estimation. These problems have resulted in considerable research and development of techniques targeting dynamic environments~\cite{Cadena2016,Saputra2018b}.

Dynamic scenes are handled by dividing the scene content into static and dynamic components in a  variety of ways. Dynamic objects can be detected using temporal methods like dense scene flows, clustering 3D motion, epipolar constraints, moving consistency checks, by tracking changes in a static map or by using instantaneous methods such as semantic segmentation. Once detected, most approaches proceed by explicitly discarding the inputs associated with the dynamic objects as outliers in pose estimation. However, simply discarding information may fail if the dynamic portions of the image are significant occluders or tend to dominate the image in terms of feature space. Hence, extracting static features is imperative to estimating the pose of the robot accurately. Most algorithms currently require an initialization phase where they map the static landmarks and world before being able to detect and track dynamic features~\cite{Bescos2018a,Tan2013b,Kim2016a}. 

In this paper, we propose a solution to reconstruct the static scene occluded by the dynamic objects in a single time step using just semantic information of the scene and light fields as sensing modality. While light fields have seen considerable use, primarily in the computer graphics community, the applicability and usefulness of light field imaging systems for mobile robotics is far more recent~\cite{Dong2013} and~\cite{Bajpayee2019}. Arrays of cameras can be used as a light field imaging system~\cite{Levoy1996}. Cameras stand out for being robust and inexpensive with a rich history in terms of their geometric and radiometric characterization.They are also relatively easy to incorporate on most robotics systems. 

A light field array captures spatial and angular radiance information at a point in space. This corresponds to a pencil of rays originating at that point and propagating in all directions as opposed to a single ray in a monocular camera. The redundancy in the pencil of rays provides information that can help us to extract the rays emerging from partially occluded portions of the scene and render synthetic aperture and digitally refocused images. In our method the static background of the scene is reconstructed not just by discarding the dynamic objects, but seeing through them via synthetic aperture refocusing. The dynamic objects are detected via deep learning based semantic segmentation. Since people are the most commonly seen dynamic class we show the results of our algorithm by detecting people in a scene. We pose the problem as a probabilistic graphical model where we jointly estimate the depth map of the static scene a well as get a refocused image of the static background using a semantic segmentation prior. We use expectation maximization (EM)~\cite{bishop2006pattern} to refine the segmentation prior for consistent labeling of dynamic objects instead of just using the predictions from deep learning. In addition, our algorithm is completely parallelizable where each pixel can be processed independently. Our method is implemented on a GPU and is capable of running close to real-time. 




\section{RELATED WORK}


Our method lies at the intersection of 3D reconstruction and light field rendering. In this section we discuss some of the recent advances in these areas that are closely related.

\textbf{\textit{Static Background Reconstruction:}} Most of the dense mapping algorithms perform static background reconstruction. \cite{Kim2016a} estimates a background model by accumulating warped depth between consecutive RGB-D images and applies energy-based dense visual odometry for motion estimation. StaticFusion~\cite{Scona2018a} is another remarkable algorithm which not only detects moving objects, but also fuses temporally consistent data instead of discarding them. A recent approach to detect the dynamic content is by semantic segmentation using deep neural networks. Including semantic information along with geometric constraints enables detecting not just dynamic objects but also potentially movable objects~\cite{Yu2018b,Bescos2018a,Barsan2018a,palazzolo2019refusion}.

All these methods require an initialization phase to build a static map from temporal data over multiple frames and then reconstruct static background in the subsequent time steps by tracking changes. All these methods use RGB-D sensor where they already have a depth map per frame. Our goal is to compute a single dense depth map of the static background within a single processing step. 

\textbf{\textit{Light-Fields for Robotics:}} Inspired by the two plane parameterization proposed by~\cite{Levoy1996},  light fields are a popular topic in computer vision and graphics for refocusing and rendering ~\cite{Isaksen2000,Lumsdaine2010,Davis2012}, super-resolution~\cite{Bishop2012} and depth estimation~\cite{Vaish2006b,Wanner2012,Wang2015,Williem2018b,Sheng2018a}. Application of light fields in robotics is not as developed but has been discussed in~\cite{Dong2013,Dansereau2011,Bajpayee2019}. Most of the robotics related work aims to solve the visual odometry problem~\cite{Dansereau2011}, and to compensate for challenging light and weather conditions~\cite{Skinner2016,Bajpayee2019,Zeller2018,Eisele2019}. None of this work addresses the problem of dynamic environments. We perform semantic guided refocusing and reconstruction to deal with dynamic objects. 

Refocusing~\cite{Isaksen2000} is performed by blending rays emerging from a focal surface and passing through a synthetic aperture without taking into account the semantics of the rays. The light field depth reconstruction techniques typically are focused on increasing the accuracy of reconstruction compromising on speed and also do not incorporate semantics. Lastly, most of the light-field research is targeted towards micro lenslet cameras~\cite{Ng2005} which suffers from small baseline separation, but for the purpose of seeing through the dynamic objects wide-baseline arrays are more suitable. Hence, we use a linear array of five cameras similar to~\cite{Bajpayee2019}, a geometry that is most suitable for deploying on  real robots.

\textbf{\textit{Multi-view Reconstruction:}} Stereo reconstruction is a well studied research area with recent methods that demonstrate real-time performance. In ELAS~\cite{geiger2010efficient}, piece-wise planar prior formed from a sparse set of matches are used to efficiently sample disparities to achieve fast computation. Inspired by this approach we also use the planar prior, but adapt the reconstruction problem to a multi-view setup incorporating scene semantics. Another important aspect of our approach is to enhance the segmentation labels produced by deep learning models so that the labels are consistent and align with 3D structure. We believe that semantic segmentation and depth reconstruction benefit from each other as shown by~\cite{kundu2014joint} and~\cite{hane2013joint} where they jointly solve the segmentation and 3D reconstruction problems. Although these methods provide promising solutions, they do not use semantic information to mask out specific objects and are not meant for practical applications with real-time constraints. Next, We propose a solution addressing these issues.


\section{LIGHT FIELD RECONSTRUCTION}

This section describes our approach to light field based static scene reconstruction for dynamic environments. 

\begin{figure}
\centering
\includegraphics[scale=0.25, clip, trim={0 2cm 0 2cm}]{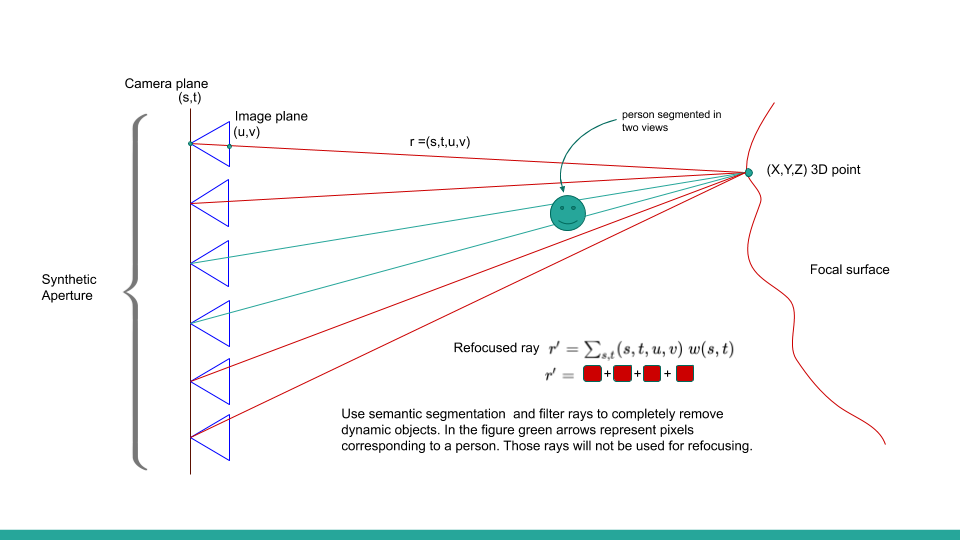}
\caption{Semantic refocusing using a light field array. Given a 3D point on the focal surface the rays projected into other cameras that correspond to static objects based on segmentation are combined to create a single refocused image.}
\label{fig:synaperture}
\end{figure}
\subsection{Problem Setup}
Suppose we have a linear array of K cameras forming images $I_k$, calibrated for both intrinsic and extrinsic parameters. This setup can be seen as a single system with synthetic aperture where each camera contributes to a ray passing through the aperture. Following the two-plane parameterization of light field~\cite{Isaksen2000}, the camera positions on the array as (s,t) coordinates lie on the entrance/camera plane and the pixels on the image plane are (u,v) coordinates, forming a unique 4D ray (s,t,u,v) as shown in \cref{fig:synaperture}. Thus, given an arbitrary focal surface F and a 3D point (X,Y,Z) on it, we can get all the rays emitting from that point and intersecting the cameras by applying the calibrated extrinsic parameters and a projective mapping combined with camera intrinsics. These rays form our synthetic aperture and can be combined via the weighted sum of the sample weights w(s,t) to compute a single refocused ray. The size of synthetic aperture represents the angular spread of the rays and  depends on the separation between the cameras on the array.   

Synthetic aperture reconstruction helps to simulate varying focus and depth of field. In free-space, all the camera rays correspond to the same point on the focal surface. But, in case of occlusions, some rays will be obstructed. A large 
synthetic aperture provides angular spread and helps us to see-through foreground objects. Choosing a focal surface on behind the foreground occluder brings the background into focus and causes the occluder to blur. This was applied by~\cite{Bajpayee2019} to see through rain and snow by manually selecting a focal plane on the road signs for autonomous navigation. 

Given the depth map of the static background $\disparity^{ref}$ and pixel coordinates $\referencePixel$ in the reference view, and the calibration parameters of the cameras array, then the rays $\pixel^k$ in the other cameras $k$ can be sampled using:  
\begin{equation}
x^k = \pi_k[R_k|t_k]\pi_{ref}^{-1}(x^{ref}, d^{ref}) \label{eq:1}
\end{equation}

where $\pi_k$ is a projective mapping between the a 3D point in space and 2D pixel coordinates on the image plane, and $R_k$ and $t_k$ are the rotation and translation of the $k^{th}$ camera. These rays are typically combined by applying an average filter~\cite{Lumsdaine2010,Isaksen2000} giving equal weight to all the rays, causing a foreground blur. Instead of using equal weights to all the rays we design a filter which selects only the rays that map to static pixels in the respective views obtained from semantic segmentation. This way, only the rays corresponding to the static background are considered and the dynamic objects in the foreground are completely eliminated. Denoting whether a pixel $\pixel^k$ belongs to the static scene or the dynamic object with $s^k$, the refocused image $I^*$ can be calculated as follows:

\begin{equation}
I^s = \frac{\sum I(x^k) * (s^k = static)}{\sum (s^k = static)}\label{eq:2}
\end{equation}

As we can observe in~\cref{eq:1}, the depth map of the static background is required to compute the refocused image. Previous work in this regard either required manual selection~\cite{Bajpayee2019} or pre-computation~\cite{Bescos2018a} of the static background, both of which are not suitable for real time applications. In this paper we calculate the depth map of the static scene and the refocused image simultaneously without any prior knowledge of the static background. To do so, we frame our problem as a probabilistic graphical model and perform MAP estimation. We further show that we can improve the semantic segmentation maps by modeling them as hidden variables within our probabilistic framework and using EM as solution method.



\subsection{Probabilistic Graphical Model} \label{sc:graphical model}
\begin{figure}
\centering
\includegraphics[scale=0.35, trim={0 0cm 0 0cm}]{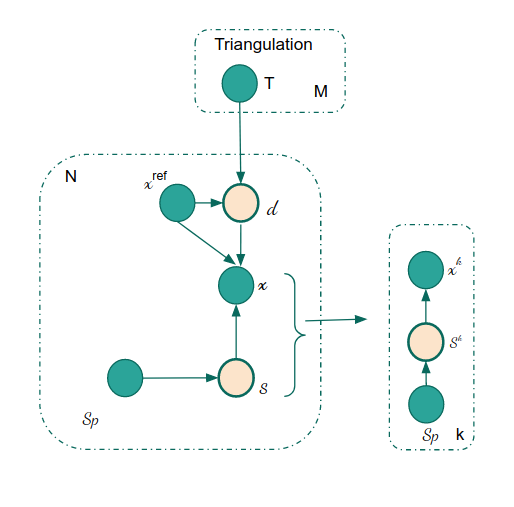}
\caption{The probabilistic graphical model. The known variables are colored solid green ($x^ref, x,\segmentationPrior, T$) and the variables to be estimated are colored white ($d,S$). The Triangulation $T$ and segmentation prior $\segmentationPrior$ are computed apriori. $d$ is disparity of a pixel in reference view, $x$ is a set of K reprojections into other cameras.}
\label{fig:pgm}
\end{figure}

The main challenge in the depth estimation of the static background is that some parts of the scene are occluded by dynamic objects (people) in some views but may be visible in others. Given an array of cameras acting as the source of image data, we need to estimate the depth map $D^{*}$ and refocused image $I^*$ of static background for a reference camera view $X^{ref}$. The key to finding correct depth is to choose the subset of camera views which exclude the pixels corresponding to dynamic objects while computing the image correspondences. We utilise scene semantics to determine static and dynamic pixels. Each camera image is used to compute per-pixel semantic labels based on a deep learning model. These semantic labels might not always be perfect and consistent across camera views due to illumination factors, photo-metric properties of the scene and occlusions. The CNN model also assigns a probability to pixels being segmented as dynamic or static: $\{ \segmentationPrior \in \mathbb{R}^k | 0 \leq \segmentationPrior^k \leq 1 \}$, where $\segmentationPrior^k = 0$ means \textit{dynamic} in $k^th$ view. Instead of using the segmentation labels as ground truth the probabilities assigned to pixels are used as a prior. We introduce a set of binary random variables $S^k$ assigned to each pixel \textit{i} in camera \textit{k} which take values either 0 or 1 representing dynamic or static pixel. These variables are conditioned on the segmentation prior $\segmentationPrior$ obtained from CNN and will be inferred from the graphical model.


The probabilistic graphical model is shown in the \cref{fig:pgm}. In this model the disparity $\disparity$ can be obtained by maximizing the posterior probability $p(d|\pixels,\referencePixel,T,Sp)$ (MAP estimate). 
The joint probability from the graphical model is computed as follows (where the normalization constant can be safely ignored during maximization):
\begin{align*}
d^* &= argmax_d \;\; p(d|\pixels,\allPriors), \\[1em]
p(d|\pixels,\allPriors) &\propto p(\disparity,\pixels | \allPriors) \\
&=  \sum_s p(\disparity, \segmentation, \pixels | \allPriors
\end{align*}
In the above equation latent variables S are introduced into the joint distribution. Expectation-Maximization (EM~\cite{bishop2006pattern}) tackles the issue of optimization with hidden (latent) variables with an iterative approach of an 'E-step' and 'M-step'. In general, assuming data $X$, the 'E-step' computes the distribution over the latent variables $Z$ according to an estimate of the parameters $\theta^{old}$: $\generalEDistr$. During the 'M-step', the estimates for Z are used to update the parameters $\theta$ by optimizing $Q(\theta) = \sum_z \generalEM)$. The trivial extension to compute the MAP estimate is to add a prior term $\generalPDistr$.
    
In our model the pixel disparities are the parameters to be optimized, and the segmentation labels are latent variables. This results in the following specification for the distributions:

\begin{align}
    &p(Z|X,\theta^{old}) &\rightarrow &\EDistr \label{eq:e-step} \\
    &\log p(X,Z|\theta) &\rightarrow &\MDistr \label{eq:m-step 1} \\
    &\log p(\theta) &\rightarrow &\log \disparityDistribution \label{eq:m-step 2}
\end{align}
Which are applied in EM as follows:

\begin{enumerate}
    \item $\segmentation \leftarrow \argmax$~\cref{eq:e-step} \label{item:e-step}
    \item $\disparity \leftarrow \argmax \sum_S$~\cref{eq:m-step 1} $\times $~\cref{eq:m-step 2} \label{item:m-step}
    \item go to (1) with $\disparity^{old} \leftarrow \disparity$
\end{enumerate}
Since it is more natural to pick an initial segmentation assignment (according to the prior) rather than a depth-estimate, we discuss (and apply) the M-step first.

\subsection{Disparity Estimation (M-step)}

Here we are interested in optimizing~\cref{item:m-step}, given a current estimate for the hidden variables $\segmentation$ given by the E-step. We assume a hard assignment for the hidden variables in E-step, which means that there is only one configuration of $\segmentation$ with non-zero probability. 
Consequently, the summation over $\segmentation$ in $Q(\disparity)$ \cref{item:m-step} collapses into a single term $\MDistr$, which corresponds to the complete log likelihood of the data and latent variables. According to the graphical model~\cref{fig:pgm}, this factorizes into the segmentation prior and pixel likelihood:

\begin{equation*}
    \MDistr \propto \log \bigg[ \pixelLikelihood + \segmentationProbability \bigg]
\end{equation*}

$\segmentationPrior$ is irrelevant during the optimization (thanks to the collapse to a single $\segmentation$ assignment). As a result, the M-step reduces to picking the disparity that optimizes:

\begin{equation} \label{eq:mstep}
    \argmax_\disparity \; \log\underbrace{\pixelLikelihood}_{\text{pixel likelihood}} + \log\underbrace{\disparityDistribution}_{\text{disparity prior}}
\end{equation}

We discuss these two factors in the following paragraphs.
\begin{figure*}[tb]
\centering
\includegraphics[width=0.31\linewidth]{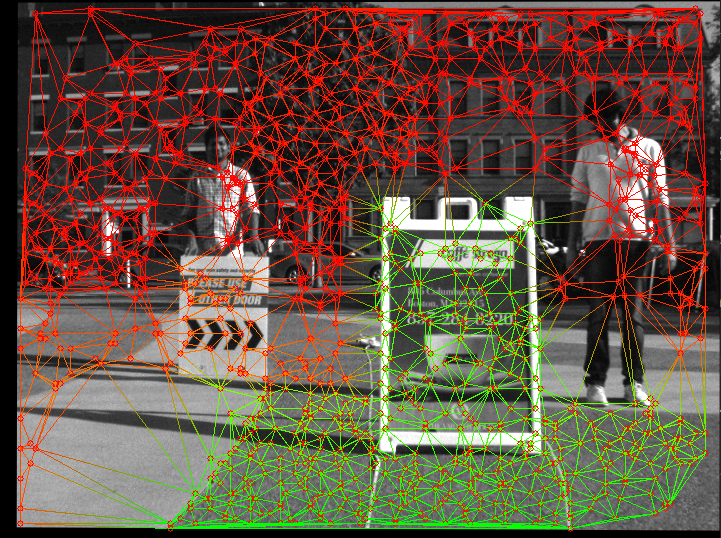}
\includegraphics[width=0.31\linewidth]{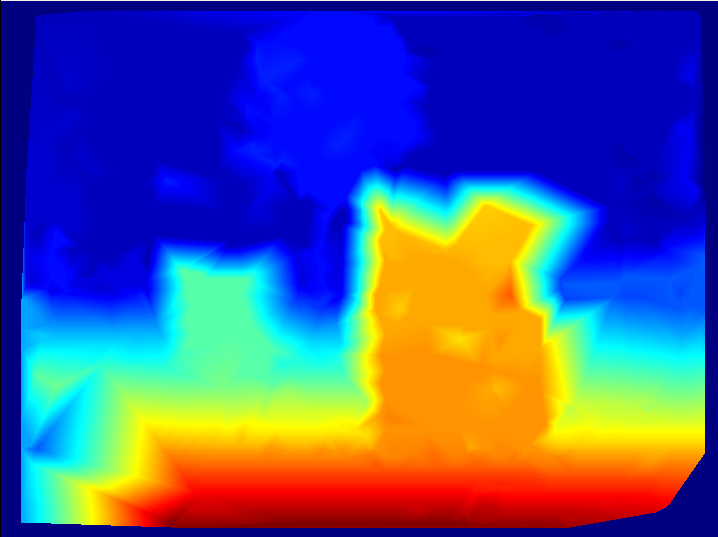}
\includegraphics[width=0.31\linewidth]{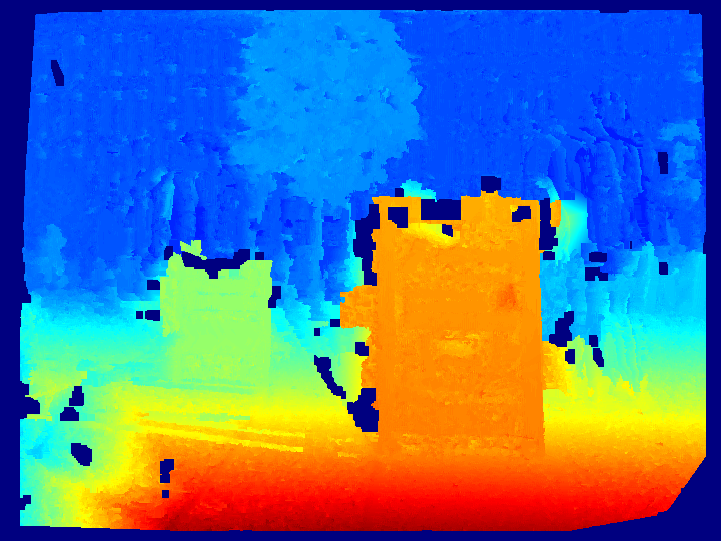}
\caption{Various steps involved in depth estimation (M-step) of the algorithm.  \textit{left:} Triangulation of the sparse set of unique support points colored based on their disparities (white: close, black : Far).  \textit{center:}The coarse disparity map formed by the piece-wise planar prior.  \textit{right:} The final refined disparity map of the static background.}
\label{fig:depth}
\end{figure*}
\subsubsection{Disparity Prior}
We use a piece-wise planar prior over the disparity space $\disparityDistribution$ by forming triangulation on a sparse set of points similar to ELAS~\cite{geiger2010efficient}.The advantage of this prior is that it helps with poorly textured regions and gives a coarse disparity map, thus reducing the search space during optimization.

In ELAS, first a sparse set of unique points are detected and matched along the full range of disparities on the epipolar lines. These support points, together with their disparities, are used to form delaunay triangulation. Unfortunately, some of the triangulation would consider the dynamic objects, so in our algorithm we exploit the knowledge from the segmentation to filter out the support points that lie on the dynamic objects. Some portions of the static background that is occluded by foreground dynamic objects might be observed from other views. So we detect support points in other camera views, compute the disparity with respect to their neighbouring camera and choose the points that re-project on to the occluded portions of the reference image. We then filter out duplicates and inconsistent support points based on their disparity values. The final set of support points are used as vertices for the delaunay triangulation. Specifically, the prior on disparity $\disparityDistribution$ is a combination of a uniform distribution and a sampled Gaussian centered about the interpolated disparity from the triangulation, for support points in the neighbourhood. 

\subsubsection{Pixel Likelihood}
The triangulation prior provides a coarse map of the interpolated disparities and a set of candidate disparities which are here used for accurate estimates based on the pixel likelihood. Given the coordinate $\referencePixel$ in the reference frame and a candidate disparity $d$, the corresponding coordinates of the source images $x^k$ can be determined using a warping function $\mathcal{W}_k(x^{ref}, d)$. This warping function represents a homography which can be computed from the reference and $k^{th}$ camera matrices~\cite{szeliski1999stereo}. This allows us to use generalized disparity space suitable for multi-view configuration as opposed to a classic multi-baseline setup where cameras perfectly placed in a plane perpendicular to their optical axes. We work in the disparity space as opposed to depth space as it facilitates discrete optimization resulting in fast computation. The pixel likelihood relies on the fact that for correct disparity value there is a high probability that warped static pixels $x^k$ in the source images will have photo metric consistency. So, we design our likelihood  function as a Laplace distribution restricted to only the static pixels such that the variance between them is minimum. The dynamic pixels don't contribute to the likelihood as they can have arbitrary intensities. As a result, the likelihood is modelled as follows:

\begin{equation*}
p(x^1...x^K \mid d,x^{ref},S) \\
\propto \begin{cases} exp\big( -\beta\; Var(f(x^1),...,f(x^K))\big)\\
\;\;\;\;\; for \;\;  x^k = \mathcal{W}_k(x^{ref}, d)\; \\
0 \;\;\;\;\; otherwise
\end{cases}
\end{equation*}

Where the variance is computed on the feature descriptors $f(x^K)$ of the pixels that are classified as static:

\begin{equation*}
Var(...) = \frac{\sum_{k=1}^K (S^k=1)[f(x^k)-\hat{f}]^2}{\sum(S^k=1)}
\end{equation*}

with $\hat{f} = \frac{\sum_k (S^k=1)f(x^k)}{\sum_k(S^k=1)}$ as the mean of the descriptors. We use the descriptors created from 3x3 sobel filter responses similar to ELAS\cite{geiger2010efficient}.


Plugging in the disparity and pixel likelihood formulae into~\cref{eq:m-step 1,eq:m-step 2} reveals the optimization problem of the M-step:

\begin{align*}
    E(\disparity) &= \beta\; Var(f(x^1),...,f(x^k)) \nonumber \\ 
    &- log\bigg[\gamma +exp\Big( -\frac{\big[d-\mu(T,x^{ref})\big]^2}{2\sigma^2}\Big) \bigg]
\end{align*}

In practice this is solved by considering all $2^k$ configurations of $\segmentation$.

\subsection{Segmentation update (E-step)}
\begin{figure*}[tb]
\centering
\includegraphics[width=0.31\linewidth]{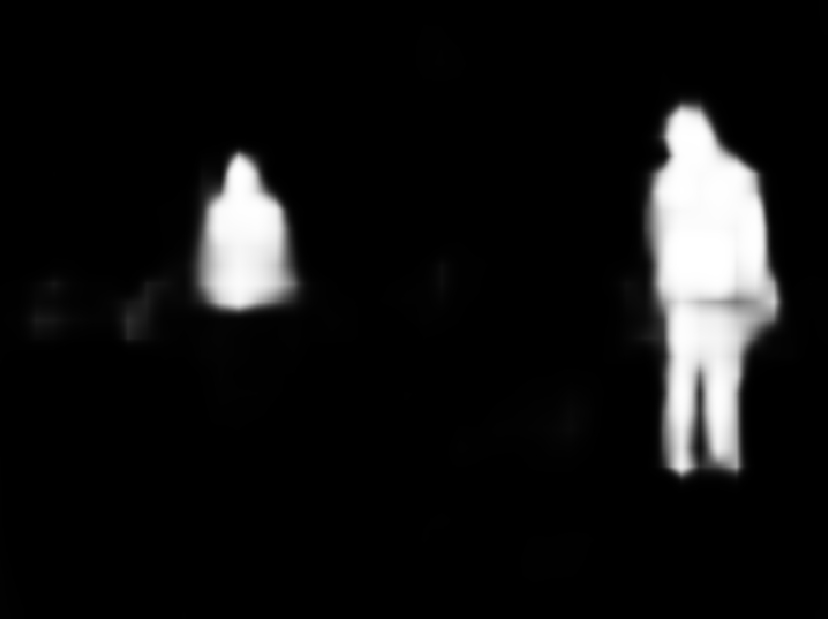}
\includegraphics[width=0.31\linewidth]{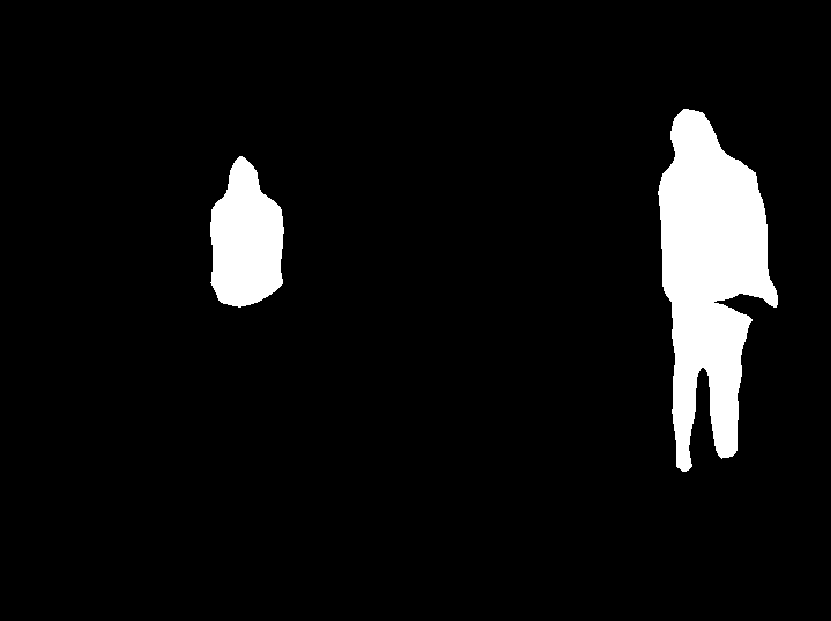}
\includegraphics[width=0.31\linewidth]{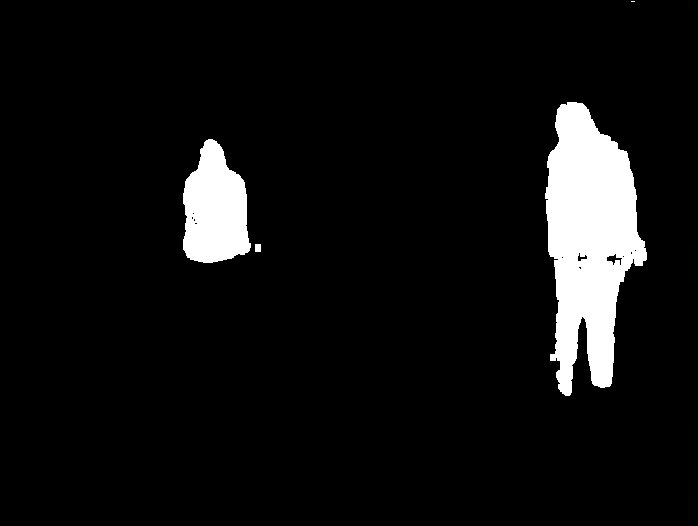}
\caption{Segmentation map update. \textit{left}: The initial segmentation prior from Bodypix.  \textit{center}: Segmentation labels obtained by simple thresholding.  \textit{right}: Final updated segmentation map.}
\label{fig:seg_update}
\end{figure*}
Assuming the segmentation algorithms is (near-) perfect, the solution to the graphical model described above is relatively straight forward. Unfortunately this proves difficult for many real-world scenarios as mentioned earlier. Modelling the segmentation as a hidden variable and the segmentation algorithm as a prior induces robustness, but comes at a cost of increased complexity of the E-step: the computation of the new segmentation distribution given the current estimate of disparity $\disparity^{old}$:

\begin{align}
p(\segmentation|\disparity^{old}, \pixels, \allPriors) &\propto p(\segmentation,\pixels|\disparity^{old},\allPriors)  \nonumber \\
&= \underbrace{p(x|\disparity^{old},\segmentation, \referencePixel)}_{\text{pixel likelihood}} \underbrace{\segmentationProbability}_{\segmentation\text{ prior}} \label{eq:e-step general}
\end{align}

The first term term, the pixel likelihood, has been discussed prior (in the M-step).
The segmentation posterior is assumed to factorize into its independent pixels, $p(\segmentation) = \prod_k p(\segmentation^K)$, where $p(\segmentation^K)$ is modelled as a delta function (e.g. 'hard assignment'). It is feasible to enumerate over these, and hence the E-step results in finding the most likely assignment of $\segmentation$: the one that maximizes (where we have substituted~\cref{eq:e-step general} with their distributions): 

\begin{equation*}
    E(\segmentation) = \beta\; Var(f(x^1),...,f(x^k)) \segmentationProbability
\end{equation*}

This results in a more accurate estimation of segmentation, which in turn helps produce a better refocused image. The number of EM iterations depends on the initial quality of the segmentation algorithm.

\subsection{Refocused Image Synthesis}

Once the EM optimization detailed above converges the estimated depth map and updated segmentation maps are used to compute the refocused image of the static background using \cref{eq:2}. We can observe that for every EM iteration where the depth map is calculated the refocused image is also computed. It usually takes 2-3 iterations for the EM to converge since we have a decent prior on segmentation, but doing EM helps us to enhance the depth and refocused image on the borders of the segmentation maps. Since the refocused image is calculated pixelwise, it tends to have specular noise due to noise in the depth image. A median filter is applied to the refocused image to get rid of these artifacts.

\subsection{Implementation details}
As mentioned earlier we restrict the classes of dynamic objects to just people and use Bodypix~\cite{bodypix} to detect and generate  pixel wise probabilities indicating the dynamic and static portions of the scene. However, our method is applicable to any moving or potentially movable objects given an appropriate segmentation. During first EM iteration $S_i^k$ values are set by thresholding the segmentation prior $\segmentationPrior >= 0.7$ as suggested~\cite{bodypix}. The depth map and refocused image of the static background are computed in a reference view. Without loss of generality we consider the left most image $X^1$ as the reference image. Thus we will have an identity transformation between $X^{ref}$ and source image $X^1$ mapping the pixel back to itself. 

During the energy minimization step we work in generalized disparity space to facilitate discrete optimization. we have an array of cameras where a unit shift in disparity between reference view and a camera may result fractional shifts with respect to another camera. Thus, while considering disparities for optimizing the energy function we make sure to include non-integral disparity values to account for multi baseline properties. The algorithm is implemented using cuda on a GPU where each pixel is independently processed for depth estimation, segmentation update to finally compute the refocused image. This results in a highly parallel real-time algorithm. Refocusing can also be performed only on the dynamic pixels in the reference frame as the other pixels already have the static background image. This further increases the speed and the computational complexity scales with respect to the amount of dynamic content in the scene.

\section{EXPERIMENTAL EVALUATION}\label{sc:experiments}
In this section we show the results of our static background reconstruction algorithm on real-world sequences collected with a custom-built light field array.

\subsection{System Setup}
\begin{figure}[tb]
\centering
\includegraphics[width=0.7\linewidth, trim={0 2cm 0 2cm}]{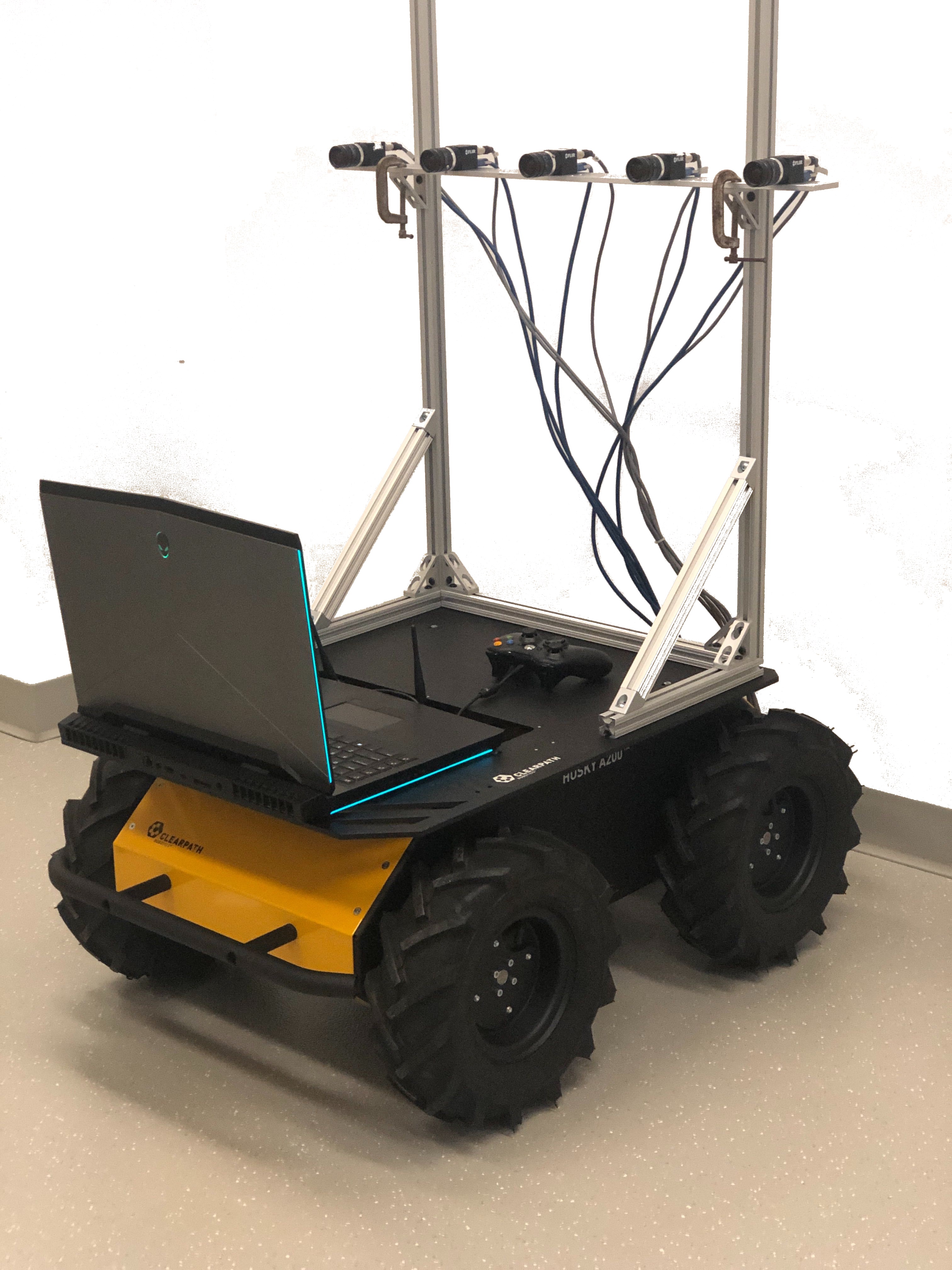}
\caption{The data collection system with the custom-built linear camera array mounted on husky.The cameras are hardware synced with a master-slave architecture.}
\label{fig:husky}
\end{figure}

\begin{figure*}[tb!]
\centering

\includegraphics[width=0.31\linewidth]{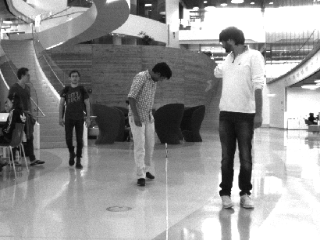}
\includegraphics[width=0.31\linewidth]{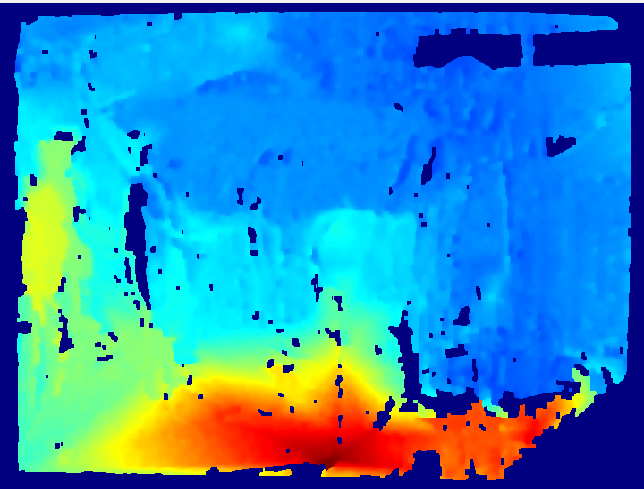}
\includegraphics[width=0.31\linewidth]{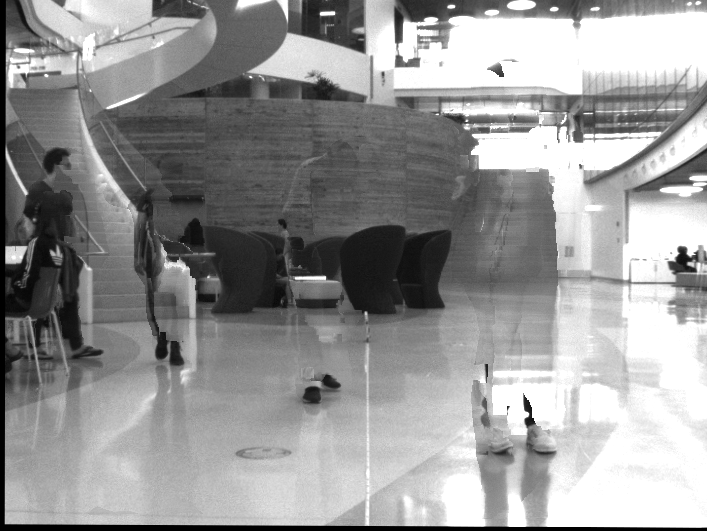}
\includegraphics[width=0.31\linewidth]{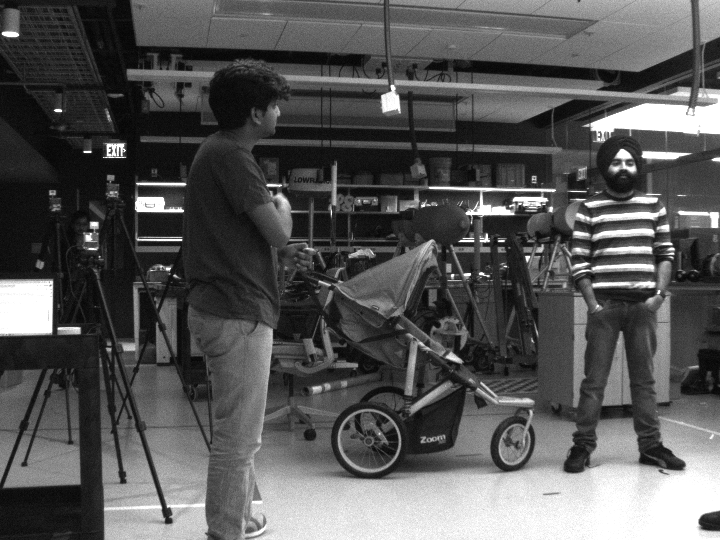}
\includegraphics[width=0.31\linewidth]{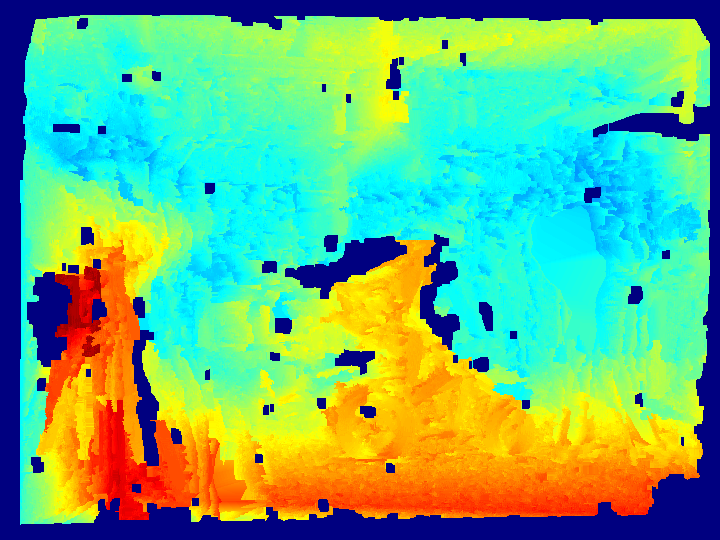}
\includegraphics[width=0.31\linewidth]{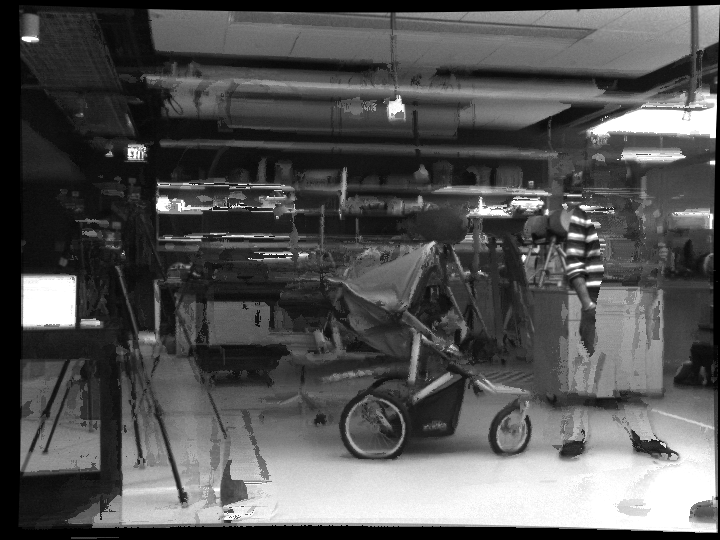}
\caption{\textit{Left:} Original images from the reference view. \textit{Center:} Disparity map of the image without people. \textit{Right:} Refocused image of the static scene estimated with our algorithm. Top row shows the result in low texture regions with plain walls. The bottom row shows the reconstruction result in present of cluttered scene.}
\label{fig:refocus_res}
\end{figure*}
Light field acquisition is done via two methods - using a large array of cameras~\cite{Levoy1996} or by using a micro lenslet array in front of image sensor~\cite{Ng2005}. Large two dimensional arrays are impractical to mount on most robots while the lenslet cameras suffer from limited parallalax due to small baseline separation. Our approach, based on constraints associated with real mobile robots uses a custom built five camera linear array that we show is sufficient for practical robotic SLAM applications. All cameras are hardware synced for synchronized image capture. The array uses a 1.6MP Pointgrey BlackflyS global shutter cameras operating at 20 fps. The camera is calibrated for camera intrinsics and extrinsics with a 9x16 checkerboard pattern using the Kalibr multi-camera calibration package\cite{furgale2013unified}. The light field data is collected by mounting the array on a Clearpath robotics' Husky UGV running ROS. All our experiments were conducted on an Microsoft Surface Pro laptop with a 6GB GEForce GTX 1060 GPU.  The algorithm runs at near real-time at ~2-3 frames per second. We can achieve faster frame rates by running the algorithm such that we are reconstructing only the image portions segmented as persons which brings the performance to ~10 fps.

\subsection{Indoor and Outdoor datasets}
We tested our approach on various datasets collected both indoors and outdoors in real-world environments with people moving in random directions. All the datasets consist of images of 720x540 pixels. We used the left most camera as the reference view for reconstruction but this choice is arbitrary and any camera can be used as the reference. We have shown the results of our static view synthesis on the outdoor dataset in \cref{fig:depth}. We also tested in scenes which has relatively more people occluding majority of the background and the algorithm does a good job of reconstructing such a complex scene.
The indoor data shown in top row of \cref{fig:refocus_res} presents a case of low textured regions and shows that the algorithm can handle such a situation quite well due to the piece-wise planar triangulation prior. In the second indoor scene we can observe the quality of the reconstruction in the presence of significant detail. In the reconstructions we can see that there are some portions of the image which still retain the original image data even though it is segmented as a dynamic object eg: feet of the person, some portions of upper body in bottom row image of \cref{fig:refocus_res}. This happens when there is not enough paralallax and none of the rays can reach the background. This is a limitation of our camera array and limitations of the baseline separation. However, as we can see in the imagery we can recover the majority of the static scene.

In \cref{fig:seg_update} the segmentation update results are shown between the original segmentation from Bodypix and the enhanced segmentation after EM. We can observe that the updated segmentation is more consistent with the 3D structure as well across  the multiple camera views which in turn helps provide an improved refocused image.


\subsection{Discussion}
There are some advantages and shortcomings that fall out of our system that are wroth pointing out.
We note that the refocused image is obtained by combining only the rays that reach the static background. If the person in the foreground is occluding the static background in such a way that no rays can reach the background, for example if the person is too close to the wall or too big for our choice of baseline separation, we will not be able to recover the background. In these cases, having temporal information could help in the reconstruction of the occluded portions of a static scene. 

Note also that shadows are not usually picked up by the segmentation algorithms but are strong candidates for key points. Even though one would detect the dynamic objects using semantic segmentation and just discard the key point associated with them, moving shadows will cause significant errors in stable localization. Our approach, however, is indifferent to shadows as we are reconstructing the static 3D scene at every time step and using this directly so that shadows do not affect us at all. 

Note that we have also arranged our cameras in a horizontal linear array to form the light field array. This naturally allows us to focus on dynamic classes that usually appear distributed vertically in the scene and for which we require horizontal parallalax to see through them.

\section{Conclusion and Future Work}
\label{sec:conclusions}
In this paper we presented a method for reconstructing the depth and 2D image of the static background from a reference view using a 4D linear light field array. We formulate this problem in a probabilistic framework where we perform an EM based optimization to estimate the depth and refocused image of the static scene and at the same time improve the semantic segmentation masks obtained from a deep learning model so that they are consistent with the 3D structure. We show promising results by evaluating our algorithm on real-world data sets collected both indoor and outdoor with people. This can be a potential front-end for SLAM to deal with dynamic environments. Through this work we have shown that light field proves to be a good candidate for robot sensing and navigation.
The main advantages of our approach are 1) We do not need an initialization phase or an initial static map for localization as we are not tracking changes in the scene. 2) Our algorithm is parallelizable and capable of running at close to real-time speed and we have demonstrated it working at 2-3 frames per second on a laptop GPU. An obvious next step would be to incorporate temporal information to get a complete dense map of the static background and We would like to further develop and demonstrate a full end-to-end light field based SLAM solution for dynamic environments.

\addtolength{\textheight}{-12cm}   



%
\section*{ACKNOWLEDGMENT}
This work was funded in part by ONR Grant N00014-19-1-2131. We would like to thank Srikumar Ramalingam, Alex Techet and Abhishek Bajpayee for helping guide our efforts and the members of the Northeastern Field Robotics lab for their help with data collection.

	
	\bibliographystyle{IEEEtran}
	\bibliography{references}
	

\end{document}